# EDGE DIRECTION MATRIXES-BASED LOCAL BINARY PATTERNS DESCRIPTOR FOR SHAPE PATTERN RECOGNITION

MOHAMMED A. TALAB[1], SITI NORUL HUDA SHEIKH ABDULLAH[2], BILAL BATAINEH

Center for Artificial Intelligence Technology,

Faculty of Information Science and Technology,

Universiti Kebangsaan Malaysia, 43600, Bangi, Selangor, Malaysia

Email: [1]mmss_ah@yahoo.com, [2]mimi@ftsm.ukm.my

## ABSTRACT

Shapes and texture image recognition usage is an essential branch of pattern recognition. It is made up of techniques that aim at extracting information from images via human knowledge and works. Local Binary Pattern (LBP) ensures encoding global and local information and scaling invariance by introducing a look-up table to reflect the uniformity structure of an object. However, edge direction matrixes (EDMS) only apply global invariant descriptor which employs first and secondary order relationships. The main idea behind this methodology is the need of improved recognition capabilities, a goal achieved by the combinative use of these descriptors.  This collaboration aims to make use of the major advantages each one presents, by simultaneously complementing each other, in order to elevate their weak points.  By using multiple classifier approaches such as random forest and multi-layer perceptron neural network, the proposed combinative descriptor are compared with the state of the art combinative methods based on  Gray-Level Co-occurrence matrix (GLCM with EDMS), LBP and moment invariant on four benchmark dataset MPEG-7 CE-Shape-1, KTH-TIPS image, Enghlishfnt and Arabic calligraphy . The experiments have shown the superiority of the introduced descriptor over the GLCM with EDMS, LBP and moment invariants and other well-known descriptor such as Scale Invariant Feature Transform from the literature.

# 1. Introduction

The simulation of numerous mathematical, statistical and heuristic approaches aimed at improving the performance of the human being is known as pattern recognition (Nieddu&Patrizi 2000). In the same way, this has given rise to a more solid and consistent study of machines and the recognition of patterns for implementation in computers. Pattern recognition has become highly significant in many fields which call for physical observation such as in the fields of remote sensing, medicine, computer vision, marketing, artificial intelligence, psychology and biology (Friedman&Kandel 1999). Image pattern recognition is a vital subject in artificial intelligence methods. The objective is to monitor, identify, label, group, mark, differentiate and understand visual patterns according to their characteristics (Liu et al. 2006). Pattern recognition involves three key phases, namely pre-processing, feature extraction and recognition. The pre-processing phase involves preparing the input image, for example by transforming the image into a scope format, removing the noise, correcting the slant, and breaking up the image into sub-images or standardizing it into an image block. The feature extraction phase involves changing the input image into characteristic images containing particular data. This data is then used in the recognition phase to categorise known or unknown patterns (Xudong 2009). According to (Xudong, 2009), feature extraction is one of the most important parts in the application of pattern recognition. He also maintained that the efficiency of each feature extraction method is closely linked to the unique resemblances in the patterns of each identical class from other patterns or noise.

Feature extraction comprises local and global methods. Local feature extraction covers the detached sections of an image, such as lines, edges, corners, shapes and sub-image areas. This procedure describes the image features that have been affected after the process of segmentation, which occurs during the pre-processing phase. Hence, the precision of the segmentation has a great influence on the local features. Joshi et al. (2007) stressed that the local method could not be used for various classes, and ultimately, the global feature extraction method involves a total or sub-regional examination of the natural image. This is normally conducted by using texture analysis techniques to extricate the global features concerning the texture of the input image. These global features are also used as common characteristics in the recognition procedure. Thus, the global method is a highly accurate technique for the categorization of more complicated datasets. Therefore, this study focused on the process of extracting of the feature in shape and texture. Previous studies indicate that a lot of researchers have focused on various processes in different ways. In this regard, this study will employ the integration of algorithms to get the best results. In spite of intricacies of the Local Binary Pattern (LBP) to ensure encoding global and local information and scaling invariance by introducing a look-up table to reflect the uniformity structure of an object. However, edge direction matrixes (EDMS) can only be applied to the global invariant descriptor which employs first and secondary order relationships. The main idea of this project is to improve the recognition capabilities. Features such as shape (Califano&Mohan 1994; Lorenz&Monagan 1995; Pentland et al. 1994) and texture (Jain et al. 1995; Pentland et al. 1994) are the main focus of this study. In this study efficient process for shape representation and texture matching will be implemented. However, the major idea is to provide compact database storage, define criteria that should be used for measuring similarity in addition to building appropriate algorithm for the shape and texture.

The objective of this study is to propose an enhance the invariant statistical feature extraction method, based on combination approach for shape and texture recognition and evaluate and compare the proposed combinatory method with the global and spatial feature extraction method with state of the art methods. This paper is organized as the following .Section 2 reviews the state of the art in the pattern recognition global and local feature extraction methods. Section 3 explains proposed shape and texture system stages and the proposed method , Section 4 present experimental and result and dataset explanation Finally, conclusions are presented in Section 5.

## 2. State of the art

### 2.1. Global Feature Extraction Approach

The global texture analysis approach has been applied in many document image analysis and recognition researches. It is based on applying texture analysis methods to text as a texture. Yong et al. (2001) have described a new texture analysis approach to handle font recognition for Chinese and English fonts. They employ the text block as a texture block, where each font has a special texture. In Other approach of feature extraction namely 2-D Gabor method was applied and tested by Ding et al. (2007), and they claimed that the relationship between the accuracy rate and the number of fonts or classes is inversely proportionate.

Ma and Doermann (2005) have proposed a new feature extraction method called the grating cell operator. It extracts the orientation texture features of the text images. This method was compared to the isotropic Gabor filter, and it classified five fonts of three scripts. The pre-processing stage of Ma and Doermann (2005) is similar to that of Yong et al. (2001). One such method was proposed by Bataineh et al. (2011) where the structure of an Arabic calligraphy font recognition system was divided into a pre-processing and a post-processing module. The texture blocks of the targeted text are produced during pre-processing, together with the removal of the edges, whereas two sub-processes are involved in the post-processing. These sub-processes are feature extraction using a statistical algorithm and recognition.

In this method have eight adjoining kernel matrices were applied and each pixel was linked to two neighbouring pixels. A connection was established between the scoped pixel, $S(x, y)$, and its neighbouring pixels, as illustrated in Figure 1 (a) and (b).

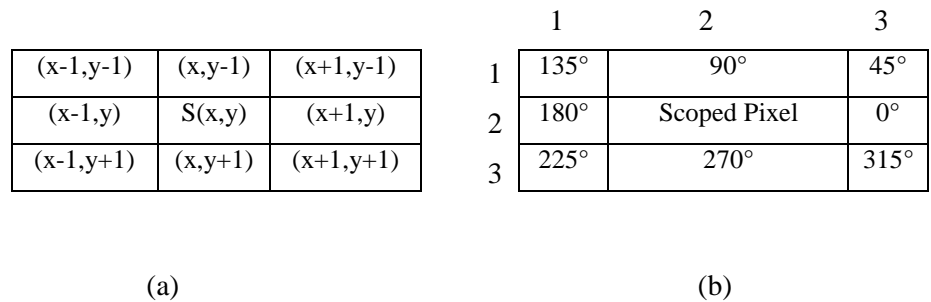

(a)          (b)

Figure 1 relationship between neighboring

The grey-level transformations, which are computed according to the displacement and angular rotation parameters, produce four grey-level co-occurrence matrices at 0, 45, 90, and 135 degrees orientation , where cells 1 & 5 are 0°, cells 2 & 6 are 135°, cells 3 & 7 are 90°, and cells 4 & 8 are 45° aligned to their nearest neighbours. An image with a spatial resolution of Nx * Ny and a grey scale level of 256 would have an angular relationship between pairs at a distance of d=1 between pixels as follows: The co-occurrence matrix would have 2NY (Nx-1) closest neighbour pairs horizontally at an orientation of 0°.

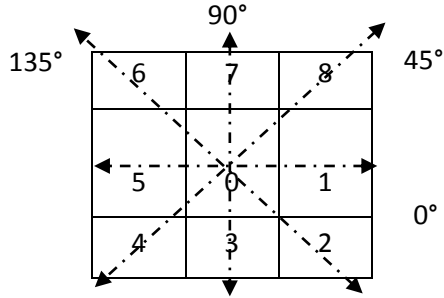

GLCM is one of the most common techniques that apply a statistical approach to global feature extraction. It consists of a matrix that assigns values that explain the distribution of occurrence in the image. The GLCM values present the number of occurrences that have grayscale value and that are related by specific relationships. The occurrence presents a pair of two pixels or two sets of pixels. If G is a GLCM matrix, I is a n _ m image and (Dx,Dy) denotes the value of the pairs of pixels that have a gray level value of i and j, as the following formula:

$$G_{\Delta x, \Delta y}(i, j) = \sum_{p=1}^{n} \sum_{q=1}^{m} \begin{cases} 1, & \text{if } I(p,q)=i \text{ and } I(p+\Delta x, q+\Delta y) \\ 0, & \text{otherwise} \end{cases}$$

..(1)

$$S(i, j) = C(i, j) + C(i, j) \qquad \ldots (2)$$

$$\text{Contrast} = \sum_{ij=0}^{N-1} P_{ij}(i-j)^2 \qquad \ldots (3)$$

$$\text{Homogeneity} = \sum_{ij=0}^{n-1} \left( \frac{P_{ij}}{1+(i-j)^2} \right) \qquad \ldots (4)$$

$$\text{Angular Second Moment} = \sum_{ij=0}^{N-1} P^2 ij \qquad \ldots (5)$$

$$\text{Entropy} = \sum_{ij=0}^{N-1} P_{ij}(-\ln P_{ij}) \qquad \ldots (6)$$

$$\text{Variance } \sigma_i^2 = \sum_{ij=1}^{n-1} p_{ij}(i-\mu_i)^2, \text{ Variance } \sigma_j^2 = \sum_{ij=0}^{N-1} p_{ij}(j-\mu_j)^2 \qquad \ldots (7)$$

$$\text{Correlation} = \sum_{i,j=0}^{N-1} P_{i,j} \left[ \frac{(i,\mu_i)(j-\mu_j)}{\sqrt{(\sigma_i^2)(\sigma_j^2)}} \right] \qquad \ldots (8)$$

Where P is the normalized matrix and N is the number of gray levels.

Various methods of feature extraction have been suggested. One such method was proposed by Bataineh et al. (2011) where the structure of an Arabic calligraphy font recognition system was divided into a pre-processing and a post-processing module. The texture blocks of the targeted text are produced during pre-processing, together with the removal of the edges, whereas two sub-processes are involved in the post-processing. These sub-processes are feature extraction using a statistical algorithm and recognition.

## 2.2. LOCAL FEATURE EXTRACTION METHOD

Local Binary Patterns (LBP) and it is used for classifying computer vision. LBP was first introduced the texture classification by Ojala et al. (1994) and it has proved to be a power full feature for texture classification. Combination of LBP and the Histogram of oriented gradients (HOG) classifier improves the detection performance significantly on some data sets. The novel descriptor is capable of improving the capabilities of a typical pattern classification system to recognize the image and it draws descriptors provided by integrating two region descriptors efficiency in recent decades. However, the feature extraction method needs to be improved in order to increase the capabilities of recognition by using formulae. This is aimed at exploring other through the completion of each other and identifies their weaknesses. The moments of most useful properties and constants related to the moment of its durability and the presence of noise is combined with global information and encoding mechanism behaviour under conditions of constant measurement, interpretation and transformation as it moves along LBP (Papakostas et al. 2012).

The Local Binary Pattern (LBP) operator is a more recent texture analysis technique that can be viewed as a statistical approach. It has been extensively accepted for both texture classification and segmentation due to its ease of use and efficiency in defining the local spatial structures of an image. Furthermore, it has been employed in a wide variety of computer vision applications.

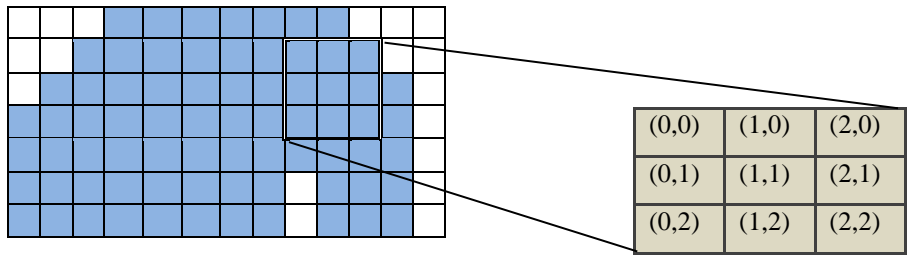

**Figure 2: find LBP for each pixel**

The other technique for extracting object's feature is Geometrical Feature (GF) topological analysis technique brought by (Arica&Yarman-Vural 2001). This technique (GF) introduces thinning, zoning contours and geometrical features (Abdullah et al. 2007; Abdullah et al. 2010).

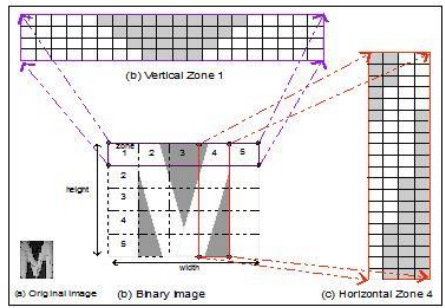

Figure 3. Images of Binary Character in GF

The image is divided into5 horizontal and 5 vertical zones which were normalized to the size of $20 \times 20$ pixels. Every character of the image is combined with horizontal zones and segmented into a $4 \times 20$ matrix and similarly, every character of the image is equally combined with vertical zones divided into a $20 \times 4$ matrix as shown in Figure 1.

The Scale Invariant Feature Transform (SIFT) algorithm is a special description proposed by (Lowe 1999). SIFT method used to find, locate and describing features in an image and robustly matches them to the images It is invariantly done by rotation, transformation and by changing the illumination of the object, Furthermore, it has an extreme distinctive sense of identifying images features that are found unique and different from that of the database. The SIFT descriptor has certain property that it uses to recognize objects that their visions are primate even though, such properties have similarity with neurons inferior temporal cortex (Lowe 2004).

Moment invariant is another local feature extraction method proposed by (Hu 1962).Hu's method had seven invariant moments.(Flusser 2000) showed that in Hu's method two rotations invariant moment among seven are dependent and five are effectively used for classification. Moreover, Flusser developed a method to find out the best sets of moments for various orders. (Wong et al. 1995) proposed a new set of invariants based on (Hu 1962) method. This improved method achieves good recognition result for a simple OCR problem using printed digits. Defines seven values, computed by normalizing central moments through order three,the"v" mean feature for moment that are invariant to object scale, position, and orientation. In terms of the central moments, the seven moments are given as:

$$v_1 = u_{20} + u_{02}$$

$$v_2 = (u_{20} - u_{02})^2 + 4u_{11}^2$$

$$v_3 = (u_{30} - 3u_{12})^2 + (3u_{12} - u_{03})^2$$

$$v_4 = (u_{30} + u_{12})^2 + (u_{21} + u_{03})^2$$

$$v_5 = (u_{30} - 3u_{12})(u_{30} + u_{12})[(u_{30} + u_{12})^2 - 3(u_{21} + u_{03})^2] \\ + (3u_{21} - u_{03})(u_{21} + u_{03})[3(u_{30} + u_{12})^2 - (u_{21} + u_{03})^2]$$

$$v_6 = (u_{20} - u_{02})[(u_{30} + u_{12})^2 - (u_{21} - u_{03})^2 \\ + 4u_{11}(u_{30} + u_{12})(u_{21} + u_{03})]$$

$$v_7 = (3u_{21} - u_{03})(u_{30} + u_{12})[(u_{30} + u_{12})^2 - 3(u_{30} + u_{12})^2] \\ + (u_{30} - 3u_{12})(u_{21} + u_{03})[3(u_{30} + u_{12})^2 - (u_{21} + u_{03})^2]$$

..(11)

## 3. The proposed method

This paper aims to solve the problem of shape and texture recognition based on statistical feature. the objective of this research work is to use improve method to extract features for the shape and texture were the introduction of images through the system in the form of pictures and then extracted bottom line binary is then determine the shape and texture entrance features extracted from this figure, so it made a lot of concessions to the shape and texture it is recognized through Multi-layer neural network with back propagation (MLBP) and the integration to get an optimal way possible, that can get the results of the input image in this research proposed method using LBP and EDMS.

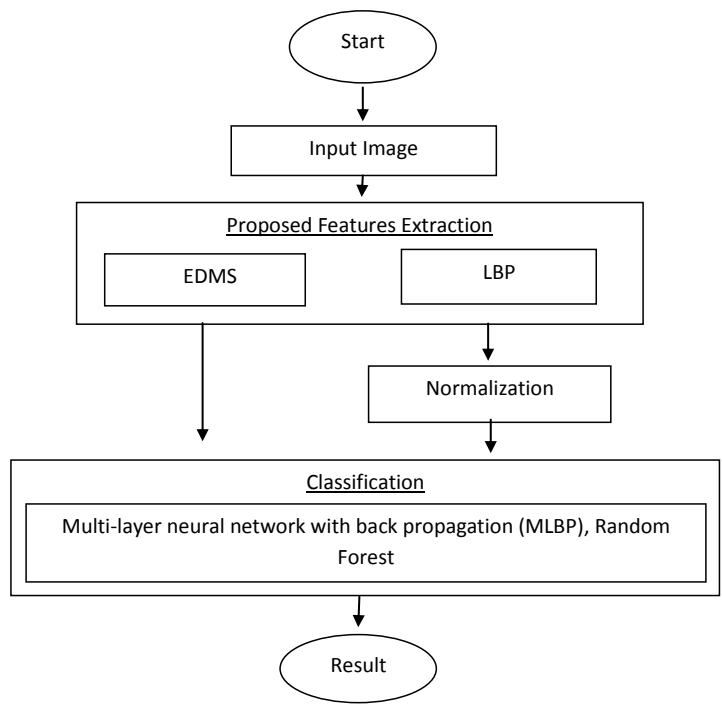

Figure 4: The proposed shape and texture recognition system flowchart.

### 3.1. The proposed Feature Extraction

The paper has two combinational types of features are used as follows:

### 3.2.1 EDMS

In the first EDMI matrix, each cell carries a position of between 0 to 315 degrees based on the pixel neighbourhood association. The relationship between the pixel values can be ascertained by calculating the occurrence of the EDM1 values, while taking into consideration the edge image of each pixel in relation to two pixels. The relationship of the scoped pixel in the edge image is represented by iedge(x, y). The following algorithm was used for calculating the EDM1:

**Algorithm for calculating the $EDM_1$**

```
For each pixel in edge
EDM1 (2, 2) = EDM1 (2, 2)  + 1.// Scoped
If  pixel (x, y + 1) = black, then // 0°
EDM1 (2, 3) = EDM1 (2, 3) + 1.
If  pixel (x + 1, y – 1) = black, then // 45°
EDM1 (3, 1) = EDM1 (3, 1) + 1.
If  pixel (x, y – 1) = black, then // 90°
EDM1 (2, 1) = EDM1 (2, 1) + 1.
If  pixel (x –1, y – 1) = black, then // 135°
EDM1 (1, 1) = EDM1 (1, 1) + 1.
If  pixel (x, y – 1) = black, then // 180°
EDM1 (1, 2) = EDM1 (2, 3) + 1.
If  pixel (x – 1, y + 1) = black, then // 225°
EDM1 (1, 3) = EDM1 ( 3, 1) + 1.
If  pixel (x, y + 1) = black, then // 270°
EDM1 (2, 3) = EDM1 (2, 1) + 1.
If  pixel (x +1, y + 1) = black, then // 315°
EDM1 (3, 3) = EDM1 (1, 1) + 1.
End
```

The second order matrix, which is 3 × 3, is regarded as an edge direction matrix and carries the relationship representation for each pixel. The most significant pixel relationship was detected by measuring the occurrence of each value in the EDM2. In instances where more than one angle had the same occurrence number, the smaller angle would be selected first, followed by the other small pixels subsequently. The algorithm for the second order relationship is given as follows:

# Algorithm for calculating the $EDM_2$

**Step 1: Sort descendingly the relationships in EDM1( x, y).**

**Step 2: For each pixel in Iedge(x, y),**

**Step 3: If Iedge(x, y) is a black pixel then**

**Step 4: Find the available relationships between two neigbouring pixels,**

**Step 5: Compare the relationship values between two available relationships,**

**Step 6: Increase number of occurrence at the related cell in EDM2(x, y).**

The results of the first-order relationship and the second-order relationships presented in EDM1and EDM2 are shown in Figure 5.

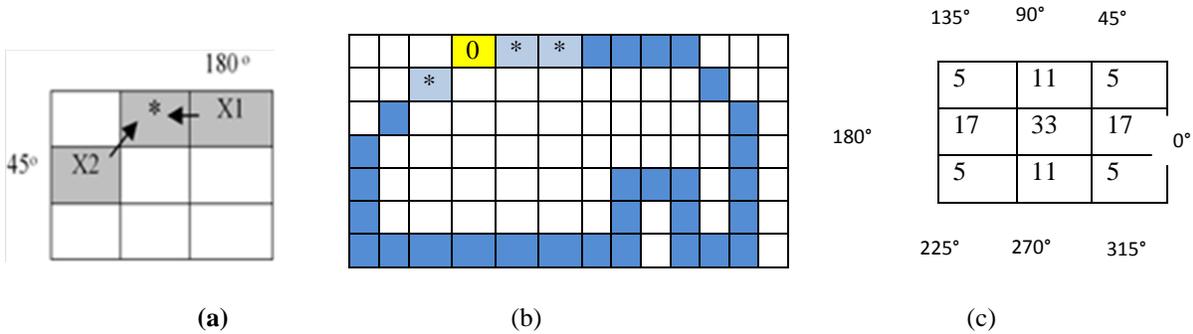

(a) (b) (c)

Figure 5: (a) The two neighbouring pixels and (b) the edge image and (c) is its EDM1.

| order | angle | value |
|---|---|---|
| 1 | 0° | 17 |
| 2 | 180° | 17 |
| 3 | 90° | 11 |
| 4 | 270° | 11 |
| 5 | 45° | 5 |
| 6 | 135° | 5 |
| 7 | 225° | 5 |
| 8 | 315° | 5 |

Table 1: The second order of the relationships.

The proposed equations extract 22 features by calculating their correlation, homogeneity, pixel

regularity, weights, edge direction and edge regularity as follows:

- Pixel Regularity: This feature denotes the percentage of distribution of the relationships in the image, and the percentage of each relationship number is calculated as follows:

$$\text{Pixel Regularity } (\theta) = EDM_1(x, y) / EDM_1(2, 2),$$

where θ represents 0°, 45°, 90° and 135°, (x, y) indicates the relative position in EDM1, and EDM1(2, 2) indicates the number of pixels in the edge image.

- Homogeneity: This feature describes the percentages of directions of the relationships and calculates the percentage of each relationship in comparison to the relationships from all angles. The following expression defines the pixel regularity:

$$\text{Homogeneity } (\theta) = EDM_1(x, y) / (\sum_{x,y=1}^{x,y=3} EDM_1(x, y)),$$

where θ represents 0°, 45°, 90° and 135°, and (x, y) indicates the relative position in EDM1.

- Correlation: This feature describes the percentage of correlations between each relationship and all other relationships. The following is the definition of the correlation expression:

$$\text{Correlation } (\theta) = EDM_1(x, y) / (\sum_{x,y=1}^{x,y=3} EDM_1(x, y)) + EDM_1(2, 2),$$

where θ represents 0°, 45°, 90° and 135°, and (x, y) indicates the relative position in EDM1.

- Weight: This feature describes the compactness of the black pixels in the image pattern by computing the percentage of the number of occurrences of each edge pixel in relation to the number of black pixels in the image source. The weight information is expressed as follows:

$$\text{Weight} = EDM_1(2, 2) / (\sum_{x,y=0}^{x,y=height,width} (Iedge(x, y) = black),$$

where only edges at 0°, 45°, 90° and 135° are taken into consideration, Iedge (x, y) denotes the original image and EDM1 (2, 2) indicates the occurrences of pixels in the image edges.

- Edge Direction: This feature describes the visual main direction by depicting the principal direction of the source image. The direction is calculated by locating a position with the maximum number of relationships.

$$\text{Edges Direction} = \text{Max}\,(EDM_1\,(x,\,y)),$$

where only 0°, 45°, 90° and 135° are taken into consideration, and (x, y) denotes the relative position in EDM1.

- Edge Regularity: : This measure indicates the directions of each pixel's relationships, which represent the percentages for all directions in EDM2 in comparison to the total number of edge pixels, as follows:

$$\text{Edges Regularity}\,(\theta^*) = EDM_2\,(x,\,y)\,/\,EDM_2(2,\,2),$$

where θ* represents 0°, 45°, 90°, 135°, 180°, 225°, 270° and 315°, and EDM2(2, 2) denotes the total number of edge pixels in the image.

### 3.2.2. LOCAL BINARY PATTERN

The Local Binary Pattern (LBP) operator is a more recent texture analysis technique that can be viewed as a statistical approach. It has been extensively accepted for both texture classification and segmentation due to its ease of use and efficiency in defining the local spatial structures of an image. Furthermore, it has been employed in a wide variety of computer vision applications. Local binary patterns (LBP) operator has been introduced by Ojala et al. [2], for texture analysis purposes .LBP is a powerful illumination invariant local descriptor, which a binary code that describes the local texture pattern is constructed by thresholding a neighbourhood of a centred pixel. If the gray value of each neighbourhood's pixel is greater than the value of the centred one, this pixel takes the value 1 in a thresholding fashion; otherwise it takes the 0 value. In this way a threshold image is constructed by applying specific weights in each pixel position so a gray-scale value in the range [0,255], which describes this small image region, is derived. By applying this operator to all available image regions of the same size, a type of histogram describing the intensity distribution of the entire image, is formed.

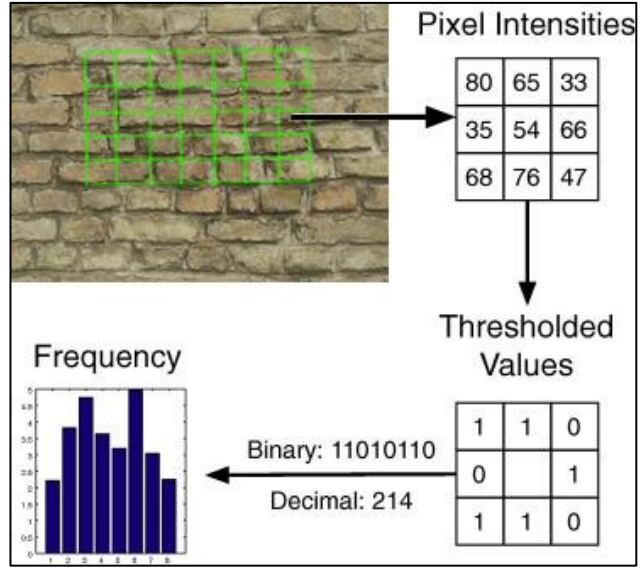

Figure 6. The LBP operation calculate of pixel

To get the LBP for pixel (1, 1) in this study, these steps are followed and then through the matrix for above figure 6 explain it.

$$f(x, y) = \begin{cases} 1, f(x, y) \geq c(x, y). \\ 0, f(x, y) < c(x, y). \end{cases}$$

In order to change the binary code to a decimal value, certain weights are assigned to the pixels belonging to the neighbourhood being processed, based on their position and by the application of the following formula:

$$LBP = \sum_{i=0}^{8-1} bi \times 2^i$$

Where $bi = \{0,1\}$ represents the binary value of the image obtained by thresholding.

### 4. EXPERIMENTS AND RESULTS

This paper described experiments conducted on this idea, and presents results and experiences that have been obtained in accordance with The proposed methods then we compare with three feature extraction for each of them and each one introduced to two classifier Multilayer Neural Network (MLNN) and Random Forest(RF). The used of different dataset MPEG-7 CE-Shape-1 dataset for shape, EnglishFnt, KTH-TIPS dataset and calligraphy Arabic for texture were compared with the SIFT, LBP-MOMENT, GLCM-EDMS in order to assess their performance based on the previous methods SIFT, GLCM-EDMS.

## 4.1 THE DATASETS

Various experiments have been conducted in this paper and various datasets have been employed for each experiment. The datasets, together with their source, description and goals. The MPEG-7 Core Experiment CE-Shape-1 is one of the most well-known benchmark datasets used in pattern recognition (Xiang Bai et al., 2010). It comprises images of binary shapes with each class producing 20 different shape images according to characteristics such as size, rotation, location, scaling and position. In figure 2 gives examples of the MPEG-7 CE-Shape-1. The english fnt dataset content of binary number and character of images for each class presenting 100 different samples from each class then contained this data on 36 class of numbers from [0 to 9] and class character from [A to Z]. The Arabic calligraphy dataset is made up of several types of images of Arabic calligraphy scripts. There are altogether 700 samples, with 100 samples from each type of image: Thuluth, Kufi, Persian, Diwani, Andalusim, Roqaa and Naskh . The dataset has 10 classes of texture materials which contains different types of images. If the number of the images in each class is counted, it will be found that each class contains 81 texture images (Zhang et al. 2007).All of dataset show in figure 7.

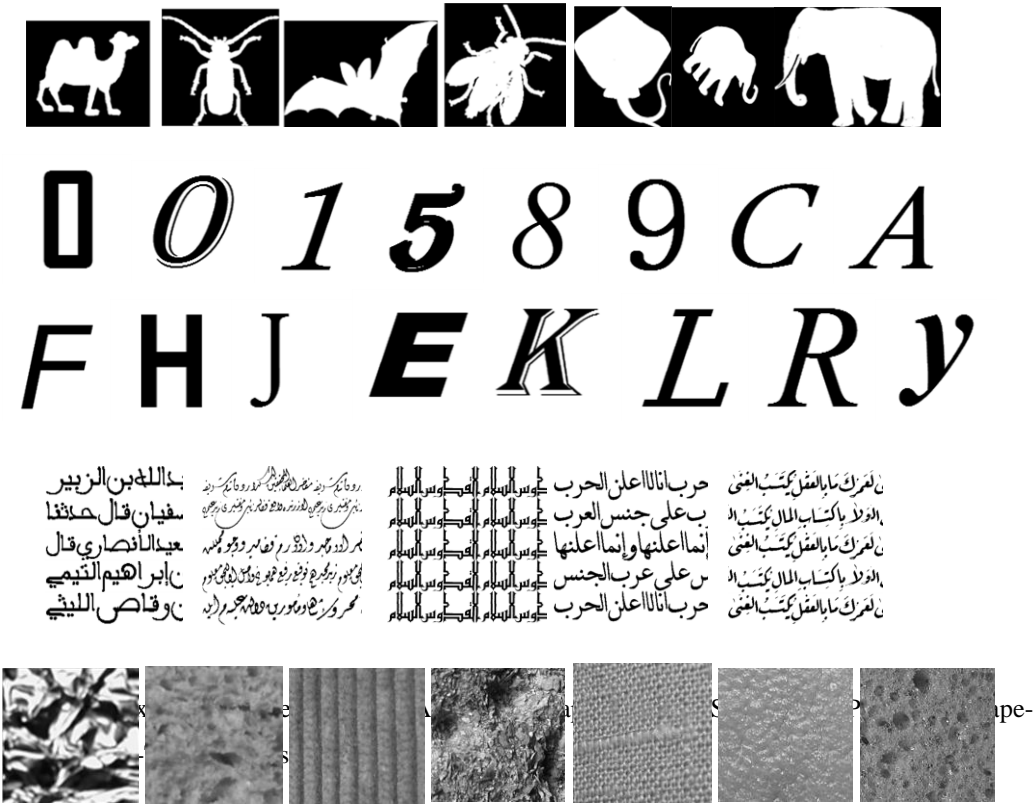

The dataset has been split into training and testing datasets. In this experiment, the training dataset is determined from percentages between 60% and 70%. Based on the experimental results, the proposed method has obtained higher accuracy rates than GLCM-EDMS and SIFT feature extraction methods in all experiments. Different percentages of training and testing data sets have been tested to determine the best performance.

| | | EXP#1 | EXP#2 | EXP#3 | EXP#4 | EXP#5 | Mean | St.Dv |
|---|---|---|---|---|---|---|---|---|
| Calligraphy Arabic | RF/GLCM-EDMS | 91.93 | 93.72 | 91.48 | 93.27 | 93.27 | 92.74 | 0.97 |
| | RF/LBP-MOMENT | 83.78 | 80.18 | 80.63 | 85.59 | 80.18 | 82.07 | 2.48 |
| | RF/SIFT | 71.18 | 70.67 | 71.11 | 71.54 | 71.90 | 71.28 | 0.47 |
| | RF/PROPOSED | 99.55 | 99.55 | 99.55 | 99.55 | 100.00 | 99.64 | 0.20 |
| | NN/GLCM-EDMS | 88.34 | 87.29 | 89.51 | 86.56 | 88.64 | 88.07 | 1.16 |
| | NN/LBP-MOMENT | 61.29 | 60.99 | 60.89 | 63.86 | 60.21 | 61.45 | 1.41 |
| | NN/SIFT | 62.35 | 62.31 | 65.27 | 61.41 | 63.81 | 63.03 | 1.52 |
| | NN/PROPOSED | 88.09 | 89.00 | 90.21 | 90.69 | 88.99 | 89.39 | 1.05 |
| ENGLISHFNT DATASET | RF/GLCM-EDMS | 76.46 | 75.91 | 76.04 | 78.69 | 77.79 | 76.98 | 1.21 |
| | RF/LBP-MOMENT | 81.95 | 81.88 | 80.84 | 80.70 | 78.82 | 80.84 | 1.27 |
| | RF/SIFT | 52.89 | 52.35 | 52.65 | 52.48 | 54.13 | 52.90 | 0.72 |
| | RF/PROPOSED | 98.26 | 98.33 | 97.63 | 98.19 | 98.47 | 98.18 | 0.32 |
| | NN/GLCM-EDMS | 40.45 | 42.51 | 45.81 | 45.95 | 43.75 | 43.69 | 2.32 |
| | NN/LBP-MOMENT | 55.85 | 60.91 | 61.89 | 57.51 | 58.71 | 58.97 | 2.46 |
| | NN/SIFT | 41.22 | 42.85 | 41.81 | 45.21 | 45.67 | 43.35 | 2.00 |
| | NN/PROPOSED | 93.01 | 93.51 | 90.64 | 90.81 | 91.31 | 91.85 | 1.32 |
| MPEG-7 CE-SHAPE-1 DATASET | RF/GLCM-EDMS | 91.84 | 92.20 | 92.91 | 92.91 | 90.49 | 92.07 | 1.00 |
| | RF/LBP-MOMENT | 88.21 | 90.00 | 86.79 | 90.00 | 88.57 | 88.71 | 1.35 |
| | RF/SIFT | 57.21 | 54.29 | 56.49 | 56.49 | 55.59 | 56.01 | 1.12 |
| | RF/PROPOSED | 98.59 | 98.23 | 99.29 | 99.65 | 97.88 | 98.73 | 0.73 |
| | NN/GLCM-EDMS | 89.36 | 90.56 | 90.76 | 88.86 | 89.21 | 89.75 | 0.86 |
| | NN/LBP-MOMENT | 77.86 | 76.51 | 78.66 | 79.34 | 78.55 | 78.18 | 1.07 |
| | NN/SIFT | 50.99 | 49.16 | 50.41 | 51.81 | 52.00 | 50.87 | 1.15 |
| | NN/PROPOSED | 95.59 | 95.81 | 96.51 | 95.23 | 94.71 | 95.57 | 0.67 |
| KTH-TIPS image dataset. | RF/GLCM-EDMS | 89.89 | 88.76 | 87.64 | 83.90 | 88.02 | 95.65 | 1.00 |
| | RF/LBP-MOMENT | 79.50 | 79.69 | 80.18 | 82.48 | 80.26 | 87.64 | 2.26 |
| | RF/GLCM-EDMS | 99.63 | 99.63 | 99.63 | 98.50 | 99.25 | 80.42 | 1.20 |
| | RF/LBP-MOMENT | 73.28 | 73.90 | 75.21 | 78.31 | 72.20 | 99.33 | 0.49 |
| | RF/SIFT | 61.05 | 60.91 | 68.00 | 65.60 | 63.69 | 74.58 | 2.35 |

| | RF/PROPOSED | 58.02 | 60.90 | 57.36 | 62.81 | 62.54 | 63.85 | 3.03 |
| --- | --- | --- | --- | --- | --- | --- | --- | --- |
| | NN/GLCM-EDMS | 80.90 | 81.40 | 85.61 | 84.51 | 80.61 | 60.32 | 2.53 |
| | NN/PROPOSED | 89.89 | 88.76 | 87.64 | 83.90 | 88.02 | 82.60 | 2.29 |

Table 2. The result for all data set using GLCM-EDMS, SIFT, LBP-MOMENT and proposed feature extraction.

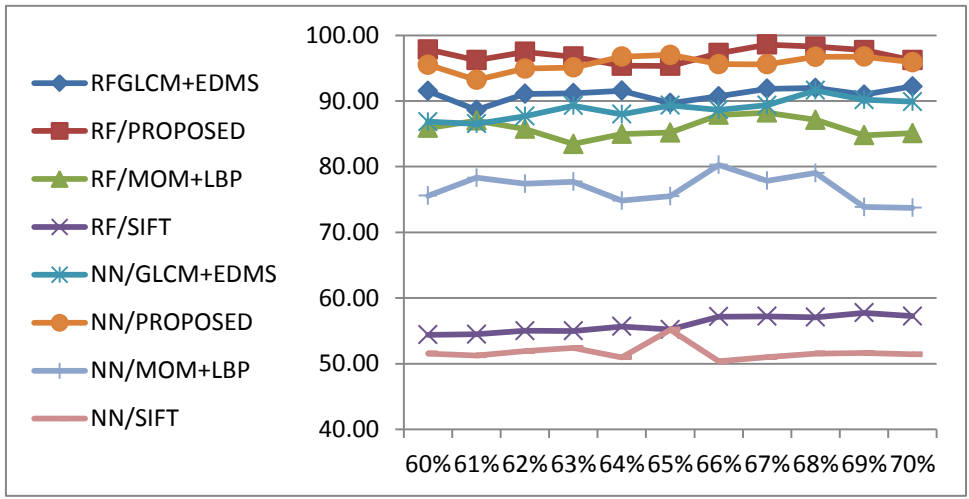

Figure 7. The classification results of the GLCM-EDMS, LBP-MOMENT, SIFT and proposed methods from 60% to 70% splitting of the training MPEG-7 CE-Shape-1 dataset and the classification results of neural network, Random Forest.

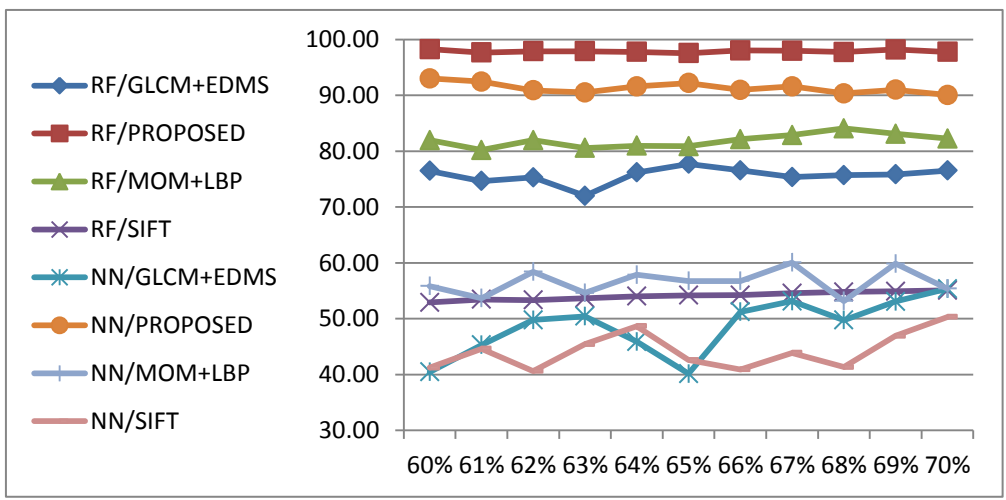

Figure 8. The classification results of the GLCM-EDMS, LBP-MOMENT,SIFT and proposed methods from 60% to 70% splitting of the training EnglishFnt dataset and the classification results of neural network , Random Forest.

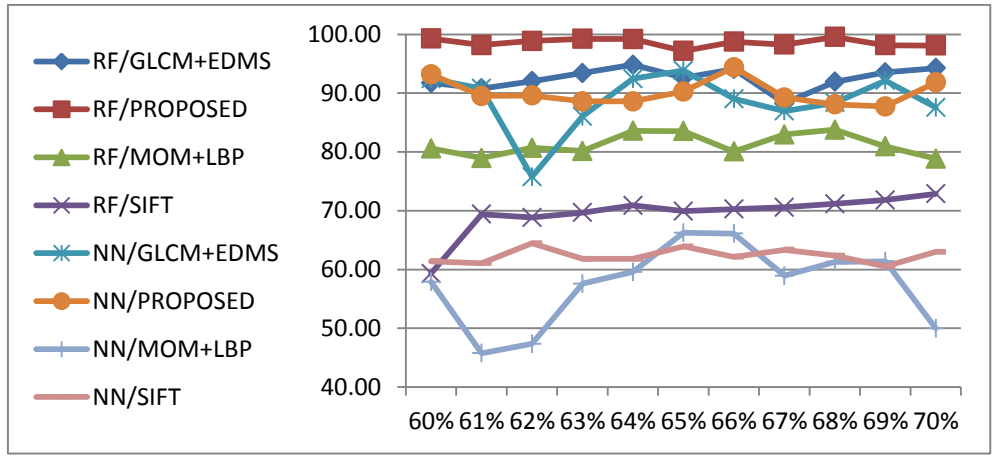

Figure 9. The classification results of the GLCM-EDMS, LBP-MOMENT, SIFT and proposed methods from 60% to 70% splitting of the training calligraphy Arabic dataset and the classification results of neural network, Random Forest.

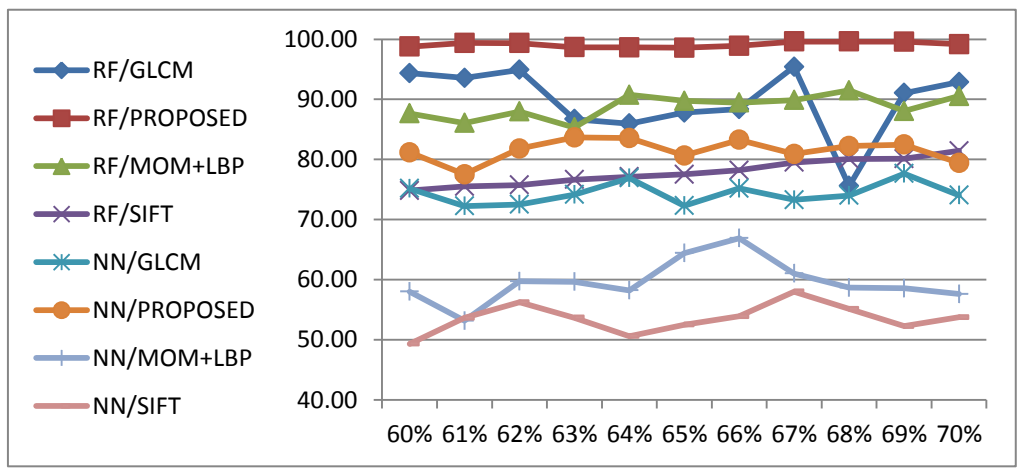

Figure 10. The classification results of the GLCM-EDMS , LBP-MOMENT , SIFT and proposed methods from 60% to 70% splitting of the training KTH-TIPS image dataset and the classification results of neural network , Random Forest .

For technical reasons, the proposed method avoided certain state-of-the-art issues. Furthermore, the proposed method does not depend on segmentation techniques. As the proposed method is based on the latest feature extraction techniques, it has been able to solve problems with a high level of accuracy compared to most state-of-the-art methods for shape and texture based on global and local feature extraction approaches In the set of experiments in this study, the results that were obtained depended on two factors: the feature extraction method that was used, and its impact on the results and classification.

The main concept behind this methodology is to improve the recognition capabilities, a goal that was achieved by using a combination of these descriptors. The aim of this integration was to take advantage of the main benefits provided by each descriptor by simultaneously complementing each other while strengthening their weak points. One such method was the EDMS-GLCM. This paper has presented the design and implementation of an improved texture analysis method for binary images called edge direction matrixes (EDMS) and local binary pattern (LBP), using a feature extraction method based on a statistical analysis of the behaviour of edge pixels in binary images.

## 5. CONCLUSION

This paper has addressed the problem of shape and texture pattern recognition which has been dealt with in previous studies on pattern recognition and texture analysis. It has proposed and implemented the statistical method of feature extraction. Combine between the local and global feature extraction approaches were presented as well as the shape and texture analysis and a comparison between the different approaches. In the feature extraction phase, the LBP and EDMS features had been proposed. In the recognition phase, we had applied two classification techniques such as the Multilayer Neural Network (MLNN) and random forest. Based on the results obtained, it had proved that the proposed method had produced the best accuracy rate, compared with the SIFT, LBP-MOMENT and GLCM-EDMS.

## Acknowledgment

We would like to express our gratitude to CAIT researcher at UKM. We also like to thank UKM University for Research Grants UKM-OUP-ICT-36-186/2010 and UKM-GGPM-ICT-119-2010 and FRGS UKM-TT-03- FRG0129-2010.

## References

Nieddu, L. & Patrizi, G. 2000. Formal Methods in Pattern Recognition: A Review. European Journal of Operational Research 120(3): 459-495.

Friedman, M. & Kandel, A. 1999. Introduction to Pattern Recognition. World Scientific.

Liu, J., Sun, J. & Wang, S. 2006. Pattern Recognition: An Overview. IJCSNS International Journal of Computer Science and Network Security 6(6): 57-61.

Xudong, J. 2009. Feature Extraction for Image Recognition and Computer Vision. Computer Science and Information Technology, 2009. ICCSIT 2009. 2nd IEEE International Conference on, hlm. 1-15.

Dhandra, B. V. & Hangarge, M.  2007.  Global and Local Features Based Handwritten Text Words and Numerals Script Identification. Conference on Computational Intelligence and Multimedia Applications, 2007. International Conference on, hlm. 471-475.

Bataineh, B., Abdullah, S. & Omar, K.  2011.  A Statistical Global Feature Extraction Method for Optical Font Recognition. Dlm. Nguyen, N., Kim, C.-G. & Janiak, A. (pnyt.). Intelligent Information and Database Systems, 6591. hlm. 257-267. Springer Berlin / Heidelberg.

Ojala, T., Pietikäinen, M. & Harwood, D.  1996.  A Comparative Study of Texture Measures with Classification Based on Featured Distributions. Pattern Recognition  29(1): 51-59.

Papakostas, G., Koulouriotis, D., Karakasis, E. & Tourassis, V.  2012.  Moment-Based Local Binary Patterns: A Novel Descriptor for Invariant Pattern Recognition Applications. Neurocomputin.

Arica, N. & Yarman-Vural, F. T.  2001.  An Overview of Character Recognition Focused on Off-Line Handwriting. Systems, Man, and Cybernetics, Part C: Applications and Reviews, IEEE Transactions on  31(2): 216-233.

Abdullah, S. N. H. S., Khalid, M., Yusof, R. & Omar, K.  2007.  License Plate Recognition Based on Geometry Features Topological Analysis and Support Vector Machine.

Abdullah, S. N. H. S., Pirahansiah, F., Khalid, M. & Omar, K.  2010.  An Evaluation of Classification Techniques Using Enhanced Geometrical Topological Feature Analysis. 2nd Malaysian Joint Conference on Artificial Intelligence (MJCAI 2010). Anjuran Malaysia, 28-30 July, 2010.

Lowe, D. G.  1999.  Object Recognition from Local Scale-Invariant Features. Computer vision, 1999. The proceedings of the seventh IEEE international conference on, hlm. 1150-1157.

Lowe, D. G.  2004.  Distinctive Image Features from Scale-Invariant Keypoints. International journal of computer vision  60(2): 91-110.

Hu, M.-K.  1962.  Visual Pattern Recognition by Moment Invariants. Information Theory, IRE Transactions on  8(2): 179-187.

Flusser, J.  2000.  On the Independence of Rotation Moment Invariants. Pattern Recognition  33(9): 1405-1410.

Wong, W.-H., Siu, W.-C. & Lam, K.-M.  1995.  Generation of Moment Invariants and Their Uses for Character Recognition. Pattern Recognition Letters  16(2): 115-123.

Xiang Bai, Xingwei Yang, Longin Jan Latecki, Wenyu Liu & Zhuowen Tu.  2010.  Learning Context-Sensitive Shape Similarity by Graph Transduction. IEEE TRANSACTIONS ON PATTERN ANALYSIS AND MACHINE INTELLIGENCE  32(5): 861-874.

Zhang, J., Marszałek, M., Lazebnik, S. & Schmid, C.  2007.  Local Features and Kernels for Classification of Texture and Object Categories: A Comprehensive Study. International journal of computer vision  73(2): 213-238